\documentclass[letterpaper, 10 pt, conference]{ieeeconf}  

\IEEEoverridecommandlockouts                              

\usepackage[backend=biber,
            hyperref=true,
            url=false,
            isbn=false,
            doi=false,
            backref=false,
            style=ieee,
            citestyle=numeric-comp,
            sorting=none,%
            block=none]{biblatex}

\DeclareNameFormat{author}{%
  \nameparts{#1}%
  \usebibmacro{name:family}{\namepartfamily}{\namepartgiven}{\namepartprefix}{\namepartsuffix}%
  \usebibmacro{name:andothers}}
\DeclareNameFormat{editor}{%
  \nameparts{#1}%
  \usebibmacro{name:family}{\namepartfamily}{\namepartgiven}{\namepartprefix}{\namepartsuffix}%
  \usebibmacro{name:andothers}}

\makeatletter
\let\NAT@parse\undefined
\makeatother

\usepackage[dvipsnames,table]{xcolor}
\usepackage{graphicx}
\usepackage{caption}
\usepackage{subcaption}
\usepackage{makecell}
\usepackage{pifont}
\usepackage{multirow}
\usepackage{amsmath}
\usepackage{arydshln}
\usepackage{booktabs}
\usepackage{nicematrix}
\usepackage{amsfonts}
\usepackage{float}
\usepackage[normalem]{ulem}
\usepackage{chngcntr}
\let\labelindent\relax
\usepackage[inline]{enumitem}
\usepackage{xspace}
\usepackage{wrapfig}
\usepackage{hyperref}
\usepackage{url}
\usepackage{arydshln}  

\definecolor{burgundy}{RGB}{144,0,32}

\newif\ifclean

\newcommand{\KGn}[1]{}

\newcommand{\jiahengn}[1]{}

\newcommand{\stat}[1]{\ifclean{#1}\else {\color{Rhodamine}{#1}}\fi}

\newcommand{\cut}[1]{\ifclean \else {\color{brown}{#1}}\fi}

\newcommand{\mashing}{mashing\xspace}

\newcommand{\Mashing}{Mashing\xspace}

\newcommand{\spoc}{$\textsc{SPOC}$\xspace}
\newcommand{\serl}{$\textsc{SERL}$\xspace}
\newcommand{\method}{$\textsc{SPARTA}$\xspace}

\newcommand{\task}{spatially-progressing\xspace}

\newcommand{\states}{\mathcal{S}}
\newcommand{\actions}{\mathcal{A}}

\newcommand{\stateinitial}{\rho_0}
\newcommand{\rewardfn}{r}
\newcommand{\transition}{\mathcal{T}}
\newcommand{\timemax}{T}
\newcommand{\policy}{\pi}

\newcommand{\custompar}[1]{
  \par
  \vspace{0.5pt}
  \noindent\textbf{#1}
}

\definecolor{LighterGray}{gray}{0.93}
\definecolor{DarkerGray}{gray}{0.73}
\newcolumntype{?}{!{\vrule width 1pt}}
\newcolumntype{g}{>{\columncolor{DarkerGray}}c}

\cleantrue  

\title{\LARGE \bf
Mash, Spread, Slice!\\Learning to Manipulate Object States via Visual Spatial Progress
}

\author{Priyanka Mandikal, Jiaheng Hu, Shivin Dass, Sagnik Majumder, Roberto Martín-Martín*, Kristen Grauman*
\thanks{The University of Texas at Austin
        {\tt\small mandikal@utexas.edu}}%
\thanks{*Equal advising}
}

\begin{document}

\maketitle


\begin{abstract}
Most robot manipulation focuses on changing the \emph{kinematic state} of objects: picking, placing, opening, or rotating them. However, a wide range of real-world manipulation tasks involve a different class of \textit{object state change}—such as mashing, spreading, or slicing—where the object's physical and visual state evolve progressively without necessarily changing its position.
We present \method, the first unified framework for the family of object state change manipulation tasks.
Our key insight is that these tasks share a common structural pattern: they involve spatially-progressing, object-centric changes that can be represented as regions transitioning from an \emph{actionable} to a \emph{transformed} state.  
Building on this insight, \method integrates \textit{spatially progressing object change segmentation maps}, a visual skill to perceive actionable vs. transformed regions for specific object state change tasks, to generate a) structured policy observations that strip away appearance variability, and b) dense rewards that capture incremental progress over time. These are  leveraged in two \method{} policy variants:
reinforcement learning
for fine-grained control without demonstrations or simulation; and greedy control for fast, lightweight deployment.
We validate \method on a real robot for three challenging tasks across \stat{10} diverse real-world objects, achieving significant improvements in training time and accuracy over sparse rewards and visual goal-conditioned baselines. 
Our results highlight progress-aware visual representations as a versatile foundation for the broader family of object state manipulation tasks.
More information at \url{https://vision.cs.utexas.edu/projects/sparta-robot} 
\end{abstract}



\vspace*{-0.25em}
\section{Introduction}
\label{sec:intro}

\begin{figure}[t]
    \centering
    \includegraphics[width=\linewidth]{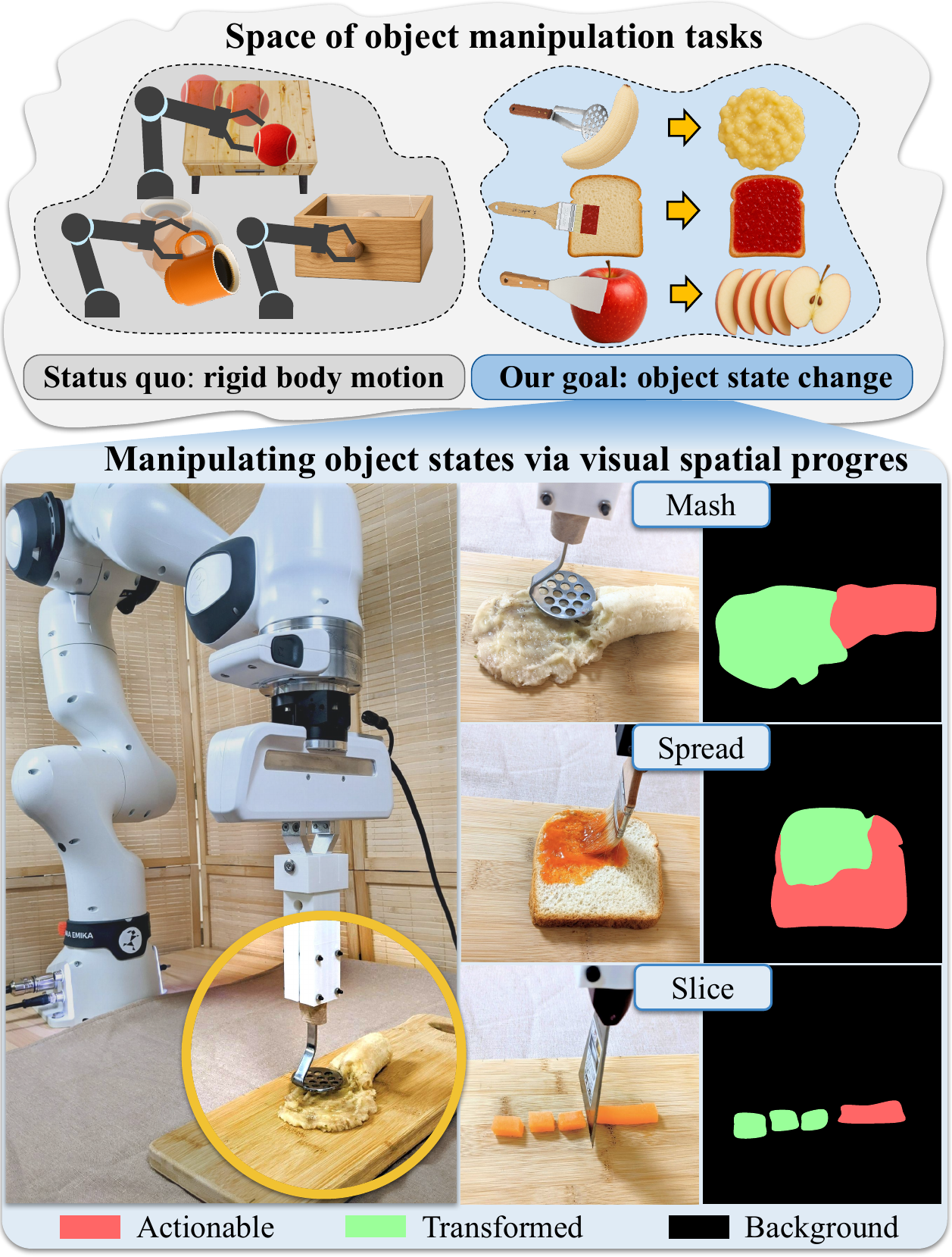}
    \caption{\small{
    \textbf{Top:} While most robotic manipulation focuses on rigid-body motion, many real-world tasks involve \textit{object state changes} such as mashing, spreading, or slicing, where objects are progressively transformed.  
    \textbf{Bottom:} SPARTA leverages spatially-progressing affordance maps of \emph{actionable} vs. \emph{transformed} regions, successfully demonstrating how to guide real robot manipulation for this family of tasks.
    }}
    \vspace{-2em}
    \label{fig:intro}
\end{figure}

The dominant paradigm in robotic manipulation centers on tasks involving rigid body motion---such as picking-and-placing~\cite{shridhar2021cliport}, opening and closing~\cite{shao2020concept2robot,bahety2024screwmimic}, pushing~\cite{pinto2017push}, or rotating~\cite{sharma2018mime} objects. While these tasks are foundational, they largely entail changing the kinematic state of objects where progress on the task is readily visible and easily monitored via changes in object pose. However, many real-world scenarios involve a fundamentally different class of manipulations: \textit{object state changes} (OSC)
\footnote{Here we adopt the term ``object state change'' (OSC) from the vision literature~\cite{souvcek2022changeit,xue2023vidosc,mandikal2025spoc}: an OSC is a transformation of an object that entails a visually distinct post-condition (e.g., chopped apple) following an action imposed on it (e.g., chopping), often with irreversible changes to the object's morphology, texture, and appearance. Not to be confused with Operational Space Control~\cite{khatib1987osccontrol}.}
~\cite{souvcek2022changeit,xue2023vidosc,grauman2022ego4d}—where an object’s physical state and visual appearance is progressively transformed, without necessarily altering its pose (see Fig.~\ref{fig:intro}, top). 
Everyday examples abound: \textit{mashing} a banana into purée, \textit{spreading} jam on bread, or \textit{slicing} a cucumber. These tasks demand continuous interaction that alters the object’s shape, texture, and color, making them both mechanically challenging and visually complex. 
Such state changes are ubiquitous in everyday activities—from cooking (e.g., grating, peeling, shredding) to household chores (e.g., painting, wiping, ironing)—yet remain largely underexplored in robotics.

What makes OSC manipulation challenging for robotics? 
Unlike motion-centric tasks, OSC requires continuous reasoning about where transformations have already occurred \textit{within} a possibly non-rigid object, where they are still needed, and how to act accordingly. Two key obstacles arise. 
First, at the \emph{representation level}, raw RGB observations entangle appearance with object state, obscuring the signals of progress and hindering generalization across objects. 
Second, at the \emph{learning level}, obtaining a good reward function is challenging: sparse success rewards provide little guidance for exploration~\cite{zhu2020ingredients}, while goal-conditioned reward functions (e.g. LIV~\cite{ma2023liv}) often rely on global scene-level embeddings that miss the \emph{fine-grained, incremental progress} essential for OSC. 
Together, these limitations make current approaches sample-inefficient and ill-suited to tasks where state changes unfold dynamically within the object.

To tackle these challenges, we propose \method (Spatial Progress-Aware Robotic object TransformAtion)—a robotic system that introduces \emph{structured, progress-aware visual affordances} tailored to OSC manipulation (see Fig.~\ref{fig:intro}, bottom). \method builds on recent computer vision advances in detecting \task object state changes (\spoc~\cite{mandikal2025spoc}), integrating them into a fully autonomous robotic system. SPOC segments a transforming object into two regions: \textit{actionable} and \textit{transformed}. 
For instance, in \mashing a potato, unmashed chunks are actionable, while mashed portions are transformed. 
\method leverages SPOC affordance\footnote{Here “affordance” means regions requiring robot interaction, distinct from conventional grasp points.} maps in two crucial ways:
(1) as structured visual observations that strip away appearance detail while preserving progress cues, enabling generalization across objects; and 
(2) as dense, spatially grounded reward signals that quantify incremental progress at each step. 
By explicitly representing “what has changed” and “what remains,” \method equips robots to reason about state progression rather than mere object kinematics.

Our formulation enables two policy variants within the same framework. In \textsc{SPARTA-L}, we use SPOC-derived rewards to train real-world RL agents from scratch—without demonstrations or simulation—achieving \textit{highly} sample-efficient learning. In parallel, \textsc{SPARTA-G} offers a non-parametric alternative, greedily steering toward nearby actionable regions in the SPOC map. 
Hence, this unified framework supports both
(1) reinforcement learning, for robust, fine-grained control in settings where noise and uncertainty demand adaptive strategies; and
(2) greedy control, for fast, lightweight deployment in simpler settings without training.
Together, \textsc{SPARTA}'s two policy variants demonstrate the versatility of its progress-aware affordances:
a single representation can power both quick heuristic controllers and data-driven RL agents, depending on task complexity.

In our experiments, we show that with just \stat{1.5--3} hours of online RL training \textit{directly in the real world} and \textit{no human demonstrations}, \method learns policies that reliably induce object state change. We evaluate across \stat{three} representative OSC tasks—spreading, mashing, and slicing—on \stat{10} diverse real-world objects, demonstrating both robustness and generality. 
By contrast, baseline methods fail to learn meaningful behavior, highlighting that dense, interpretable affordances for object state change are key to enabling sample-efficient, generalizable real-world robot learning—charting a path beyond rigid-body manipulation.

\section{Related Work}\label{sec:rel_work}

\custompar{Non-rigid object manipulation.}
Recent advances tackle individual tasks requiring more complex manipulation than traditional pick-and-place-style tasks,
such as cutting~\cite{heiden2021disect,xu2023roboninja,beltran2024sliceit,shi2023robocook}, peeling~\cite{ye2024morpheus,chen2024vegetablepeeling,foodpeeling2021dualarm}, and stir-frying~\cite{stirfry2024}. 
However, these efforts tackle each task in \textit{isolation}, often focus on the task's mechanical aspects,
lack general-purpose vision feedback,
or rely heavily on simulation. 
In contrast, our work targets a broad class
of spatially transformative tasks that require reasoning over visual state changes rather than contact dynamics alone, exploiting a \textit{unified} visual representation that is shared across objects and state-change tasks.

\custompar{Visual representations for robot learning.}
To accelerate downstream policy learning, recent works pretrain
visual representations on large-scale data~\cite{nair2022r3m,Radosavovic2022,ma2022vip,ma2023liv}. 
\cut{Early efforts used labeled Internet images~\cite{shah2021rrl} or robot-specific data~\cite{vision2action}, while recent methods pretrain on large-scale human activity videos~\cite{grauman2022ego4d,Damen2018EPICKITCHENS}.}
More relevant to our novel visual rewards,
VIP~\cite{ma2022vip}  learns 
an implicit value function over egocentric videos, while its extension LIV~\cite{ma2023liv} further incorporates language-goal embeddings.
There is also growing interest in \cut{using }LLMs~\cite{ma2023eureka} \& VLMs~\cite{rt22023corl} for robotic reasoning, typically using\cut{ coarse}
frame-level goal matching or symbolic planning.
In contrast, \method leverages a VLM for spatial reasoning over localized object regions, enabling dense reward generation and supporting both efficient planning and online RL for visually complex manipulation.

\custompar{Affordances in robotics.}
Understanding \textit{how} and \textit{where} to interact with objects has driven a surge of interest in affordance-based functional grasping~\cite{brahmbhatt2019contactgrasp,mandikal2021graff,mandikal2021dexvip,wu2022learning,agarwal2023dexterous,bahl2023affordances}.
Parallel efforts in computer vision predict hand-object interactions
~\cite{hasson19_obman,ye2023affordance,liu2022joint}, 
but they emphasize pick-and-place or grasping tasks. 
In contrast, we tackle a fundamentally different class of affordance---spatially evolving, visual object state transformations that generalize across tasks and robot embodiments. To our knowledge, this represents the first affordance reasoning approach for such manipulations achieving non-rigid object interactions on a real robot.

\custompar{Object state change understanding.}
OSC is explored in computer vision for video-level \cut{state }classification~\cite{souvcek2022changeit,xue2023vidosc}, \cut{object }segmentation~\cite{yu2023video,tokmakov2023vost,mandikal2025spoc}, and generation~\cite{soucek2024genhowto}. Our work is inspired in part by the \textit{spatially progressing object state change} (SPOC) task~\cite{mandikal2025spoc}, which segments state-changing objects into actionable and transformed regions. Trained on large-scale instructional ``how-to" videos~\cite{miech2019howto100m}, SPOC exhibits robust spatial reasoning across diverse objects and transformations. However, these models are vision-only: they passively analyze state changes but 
do not inform robot control. Our work bridges that gap.  By integrating vision-based OSC understanding into robot manipulation, we show how robots can learn to act using SPOC-style affordances capturing gradual visual progress---difficult to address with tactile sensing~\cite{foodpeeling2021dualarm}, force models, or binary state classifiers~\cite{shao2020concept2robot,oscpddl2024}.

\custompar{Real-world Reinforcement Learning} 
Real-world RL enables autonomous policy learning directly from real-world interactions, avoiding the need for explicit world models or high-fidelity simulators. This makes it particularly promising for contact-rich manipulations, where accurate modeling is notoriously difficult \cite{hu2025slac,zhang2025rewind}. 
However, when tasks require progressive object state changes, existing methods struggle on two fronts: first, learning visual representations that capture subtle intra-object changes; and second, defining reward functions that provide dense, informative feedback \cite{luo2024serl, zhu2020ingredients}. These challenges lead to poor sample efficiency and hinder real-world applicability. Our work tackles both issues by leveraging spatially progressing OSC segmentation maps, leading to successful policy learning on challenging tasks.

\section{Robotic Object State Change}
\label{sec:approach_a}

\begin{figure}
    \centering
    \includegraphics[width=\linewidth]{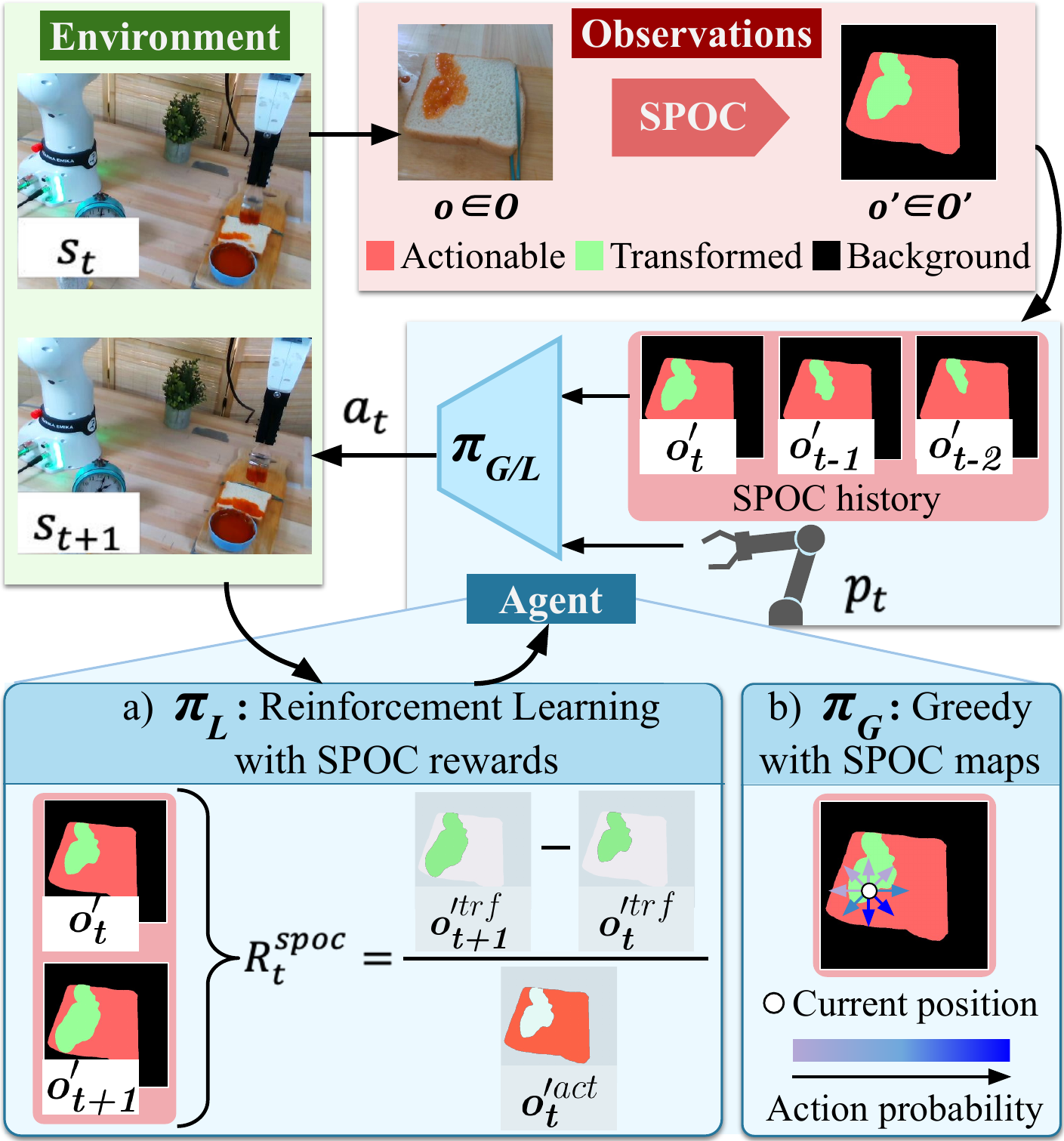}
    \caption{\small{\textbf{Overview of \method.} 
At each episode step, our policy takes the current and past SPOC~\cite{mandikal2025spoc} visual-affordance (segmentation) maps as inputs (Sec.~\ref{subsec:spoc}), along with the robot arm's proprioception data and predicts a  displacement action for the arm's end-effector. 
SPARTA supports two robot policy variants:
(a) \textbf{SPARTA-L} (Learning): a reinforcement learning agent trained using a dense reward that 
measures the progressive change of object regions from \textit{actionable} (red) to \textit{transformed} (green)
(Sec.~\ref{subsec:sparta-l});
(b) \textbf{SPARTA-G} (Greedy): selects among 8 discrete directions based on the local density of actionable pixels, producing a fast, greedy policy guided by visual progress (Sec.~\ref{subsec:sparta-g}).
}}
    \vspace*{-0.2in}
    \label{fig:model}
\end{figure}

Our goal is to enable robots to perform \emph{object state change} (OSC) tasks, where the object’s morphology, texture, or appearance evolve progressively over time. Unlike traditional robotic manipulation, which focuses on moving rigid objects in space (e.g., pick-and-place or pushing), OSC tasks demand reasoning about transformations \textit{within} the object itself. The challenge is not simply to change an object’s kinematic state, but to decide where and how to act on deformable regions so as to drive continuous, irreversible changes in the object’s physical state. This fundamentally alters the problem: the robot must perceive gradual transformations, localize actionable regions, and sequence fine-grained actions that accumulate toward a globally transformed outcome.

\custompar{Problem Formulation.} 
We formulate OSC task as a Partially Observable Markov Decision Process (POMDP) $(\states, \actions, \transition, \Omega, \rewardfn, \stateinitial, \gamma)$, where $\states$ are the true environment states, $\actions$ are robot actions, $\Omega$  are the observations, $\transition(s_{t+1} \mid s_t, a_t)$ governs state evolution, $\rewardfn(s_t, a_t)$ provides feedback, $\stateinitial$ is the distribution over initial states, and $\gamma$ is the discount factor. The goal is to learn a policy $\policy(a_t \mid \omega_{\leq t})$ that maximizes expected discounted return:
$J(\policy) = \mathbb{E}_\policy \bigg[\sum_{t=0}^{\timemax-1} \gamma^t \rewardfn(s_t, a_t)\bigg]$.

Here, partial observability arises because the underlying object state (e.g., which region of a banana is mashed) is not directly available—only visual observations and proprioception are accessible. Unlike motion-centric tasks where object poses provide a sufficient state proxy, in OSC we need observation spaces that faithfully approximate these hidden, spatially evolving states.

\custompar{Observation Space.} 
The robot operates in a tabletop workspace with a single object presented at random orientations. Each observation $\omega_t \in \Omega$ is represented by visual and proprioceptive components, $\Omega = O \times P$. At time $t$, the robot receives an RGB observation $o_t \in O$ from a fixed camera and proprioceptive input $p_t \in P$ encoding end-effector position. 
The key challenge is that raw RGB frames, while visually rich, entangle object-specific appearance with the underlying dynamics of state change. For robots, this makes it difficult to learn sample-efficient, generalizable policies from limited real-world data. What is needed instead are structured visual abstractions that strip away appearance-specific detail while preserving cues that reflect how the object is evolving over time, bringing the observation space closer to the task-relevant state representation.
We introduce such a representation in Sec.~\ref{subsec:spoc}.

\custompar{Action Space.}
While classical manipulation often requires planning global object motions, OSC tasks demand acting at \emph{specific intra-object locations} to drive localized transformations (e.g., pressing unmashed potato chunks or brushing uncoated regions of bread).
To reflect this, we constrain the action space to a 2D manifold aligned with the object surface, enabling policies to directly reason about \emph{where} to apply tool actions. Concretely, the policy outputs continuous $\Delta x, \Delta y$ displacements, sampled from a Gaussian centered at the mean predicted by $\policy$. 
At the resulting $(x,y)$ location, a task-specific primitive is executed: sweeping motions for spreading, downward pressing for mashing, or slicing strokes. This structured action space both mirrors the spatially progressive nature of OSC tasks and reduces complexity, making it possible to learn sample-efficient policies that generalize across objects.

Next we provide a detailed framework for SPARTA, leveraging visual spatial progress to solve robotic OSC tasks.

\section{SPARTA: Robot Policies for OSCs \\via Visual Spatial Progress}
\label{sec:approach_b}

\subsection{Integrating SPOC Visual Affordances for Robotics}
\label{subsec:spoc}

To instantiate structured visual abstractions for object state change manipulation, we build on the \textit{Spatially Progressing Object State Change} (SPOC) task~\cite{mandikal2025spoc} from computer vision. SPOC segments objects undergoing transformation into two regions: those that remain \textit{actionable} and those already \textit{transformed}. Given a sequence of RGB frames ${o_1, o_2, \ldots, o_T}$, the model outputs masks $o'_t=\{o'^{act}_t, o'^{trf}_t\}$ per frame (see red-green maps in Fig.~\ref{fig:model}, top right). For example, in a coating task, plain bread regions are actionable, while coated ones are transformed. Importantly, while RGB frames serve as input, the robot policy itself only receives the SPOC masks $O'$—a distilled, task-relevant abstraction that strips away appearance variability (see Fig.~\ref{fig:model}, middle). This decouples the heavy-lifting of visual affordance prediction (handled by a pretrained vision model) from policy learning, which now operates on interpretable, object-centric maps that transfer more readily across objects.

\begin{figure}[t] 
    \centering 
    \includegraphics[width=1.0\linewidth]{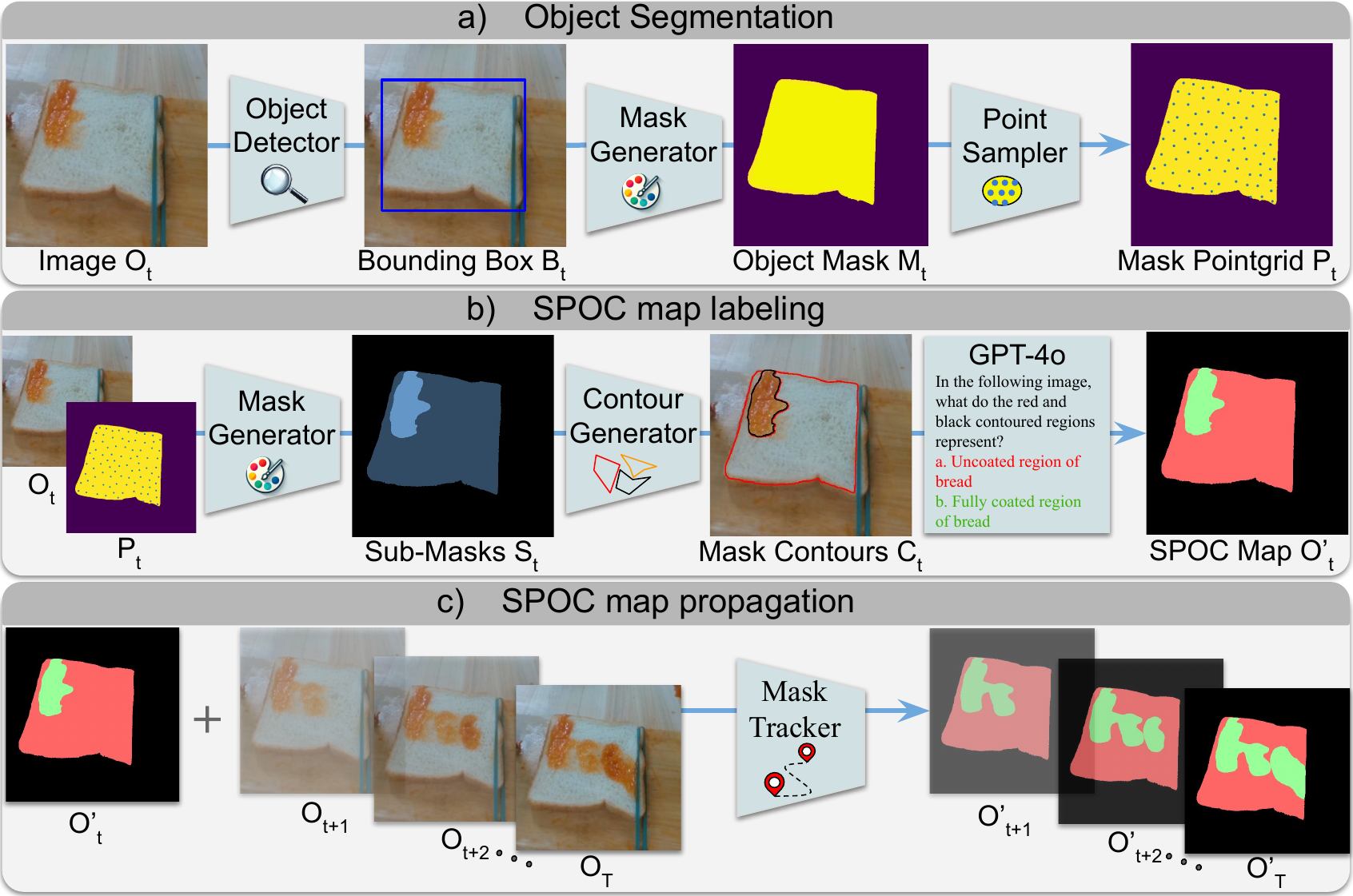} 
    \caption{\small{\textbf{Our SPOC affordance map generation pipeline.} 
    \textbf{(a)} Grounded-SAM~\cite{ren2024gsam} is used to extract an object mask from the initial frame.
    \textbf{(b)} Farthest-point sampling generates intra-object regions, classified into \textit{actionable} or \textit{transformed} by prompting GPT-4o~\cite{openai2023gpt4} using color-coded overlays.
    \textbf{(c)} Once classified, transformed regions are tracked across subsequent frames using DeAOT~\cite{yang2022deaot} to maintain temporal consistency with minimal computation.
    }}
    \vspace*{-0.2in}
    \label{fig:spoc_map_gen} 
\end{figure}

We adapt SPOC for robotics by generating SPOC affordance maps directly from real-time visual observations. While prior work~\cite{mandikal2025spoc} leverages Grounded-SAM~\cite{ren2024gsam} and CLIP~\cite{clip}, we find that replacing CLIP with a stronger vision-language model (VLM) such as GPT-4o~\cite{openai2023gpt4} significantly improves segmentation accuracy—particularly in distinguishing intra-object regions (e.g., partially mashed banana). 
Instead of assigning a single label to the entire object mask, we sample multiple intra-object regions using farthest-point prompts with SAM, and classify each via GPT-4o into actionable or transformed states. Since per-frame GPT queries are slow ($\sim$5s), we introduce a fast mask propagation strategy using DeAOT~\cite{yang2022deaot} tracking ($\sim$0.2s/frame) to boost real-time throughput for robotic control.
See Fig.~\ref{fig:spoc_map_gen} for the full pipeline.
These affordance maps offer dense, object-centric structure that is crucial for shaping progress-based rewards and guiding spatial-aware policy learning.

We call attention to two key aspects of our design.  First, by virtue of the VLM training, the SPOC masks are derived from large-scale data of \emph{human} action in natural environments. 
This allows them to encode what transformations \emph{look like} rather than how humans physically move, avoiding embodiment mismatch while still transferring visual know-how to robots.
Second, by reducing raw RGB to binary actionable/transformed segmentations, our method enables generalization across new objects (e.g., from bread to cheese) and materials (e.g., from ketchup to ranch), since the policy no longer depends on appearance but on \emph{consistent} notions of visual progress.

SPARTA leverages these affordances in two complementary ways:
\textsc{SPARTA-L} (~\ref{subsec:sparta-l}), a learning algorithm that uses SPOC rewards to drive real-world online RL; and
\textsc{SPARTA-G} (~\ref{subsec:sparta-g}), a greedy algorithm that uses SPOC maps to enable greedy action selection.
The shared MDP formulation across both variants, grounded in SPOC-based state and reward representations, underscores the versatility of this representation—enabling both adaptive policy learning and reactive greedy planning within a unified framework.

\subsection{SPARTA-L: Reinforcement Learning with SPOC rewards}
\label{subsec:sparta-l}

Object state change tasks demand \emph{sequential decision-making}: each action transforms only a local region of the object, and the robot must continually decide \emph{where to act next} to optimize long-term task success. 
Reinforcement learning (RL) naturally fits this setting, as it optimizes long-horizon returns rather than immediate feedback.

A central challenge in real-world RL, however, is reward design. Sparse, binary success signals provide little guidance for sample-efficient training, while dense, automated feedback is rarely available~\cite{zhu2020ingredients}.
For instance, coating an additional patch of bread or mashing a new section of banana is meaningful progress toward the goal, but this nuance is lost with simple binary rewards. As we show in Sec.~\ref{sec:expts}, such dense feedback is essential for stabilizing exploration and driving sample-efficient learning in the real world.

To address this, we design a dense, spatially grounded reward function that reflects incremental progress and enables \textit{real-time, demonstration-free} learning:
\setlength{\abovedisplayskip}{1em}
\setlength{\belowdisplayskip}{1em}
\begin{equation}
        \rewardfn_t = \alpha R^{spoc}_t + \beta R^{succ}_t + \eta R^{entropy}_t.
\label{eq:reward}
\end{equation}
Here, 
$R^{succ}_t$ 
is a sparse terminal reward for task completion (when $>$95\% of the object is transformed), 
and $R^{entropy}_t$
promotes action diversity and exploration. The key novel component, 
$R^{spoc}_t$ (see Fig.~\ref{fig:model}-a), 
provides dense feedback at every step by quantifying 
\textit{newly transformed} area since the previous timestep:
\setlength{\abovedisplayskip}{1em}
\setlength{\belowdisplayskip}{1em}
\begin{equation}
    R^{spoc}_t = \frac{A_{t+1}^{trf}-A_{t}^{trf}}{A_t^{act}}
\label{eq:reward_spoc}
\end{equation}
where $A_t^{trf}$ and $A_t^{act}$ denote the transformed and actionable areas of SPOC segmentation maps $o'^{trf}_t$ and $o'^{act}_t$.
Unlike sparse success rewards, this formulation rewards incremental transformation of new regions, focusing the agent on actionable areas while avoiding redundancy.

The result is an object-centric, task-agnostic reward that captures fine-grained spatial progress from vision alone. It eliminates the need for simulation, privileged state, or human demonstrations, and---as shown in Sec.~\ref{sec:expts}---yields smooth, monotonic learning curves for real-world OSC tasks.
For policy optimization, we build on
\serl~\cite{luo2024serl}, using Soft Actor-Critic (SAC)~\cite{haarnoja2018sac} with regularization from RLPD~\cite{ball2023rlpd} for sample-efficient off-policy learning directly in the real world. 
Unlike \serl, however, \method does not rely on any human demonstrations. Instead, visually grounded rewards derived from SPOC affordances are sufficient to drive sample-efficient real world RL.

\subsection{SPARTA-G: Greedy Policy with SPOC Maps}
\label{subsec:sparta-g}

While RL is a general formulation for policy learning under noisy perception, some OSC tasks can be solved effectively with a simpler controller. In particular, when using large, symmetric tools (e.g., a masher), each action covers a broad surface area, making the control policy less sensitive to errors in perception or misalignment between visual and proprioceptive inputs. In such cases, a greedy policy that simply moves toward untransformed regions can work well without training. By contrast, tasks that require precise, directional control with thin, asymmetric tools (e.g., spreading with a brush) are much more vulnerable to noise, where RL offers a clear advantage.

To capture this easier regime, we introduce \textsc{SPARTA-G}, a non-parametric greedy controller that directly exploits spatial priors in SPOC maps. 
At each timestep $t$, the agent receives a segmentation $o'_t \in \mathcal{O'}$ labeling pixels as \textit{actionable}, \textit{transformed}, or \textit{background}. 
Given the current end-effector position $p_t$, the controller evaluates a discrete set of 8 candidate motions $\{a_t^{(i)}\}_{i=1}^8$ corresponding to uniformly spaced directions in the $xy$-plane (see Fig.~\ref{fig:model}-b). 
For each candidate, we compute the density of actionable pixels in a neighborhood $\mathcal{N}(p_t + a_t^{(i)})$ of the predicted endpoint, and select the direction with the maximum density:
\begin{equation}
a_t = \arg\max_{a_t^{(i)}} \sum_{x \in \mathcal{N}(p_t + a_t^{(i)})} \mathbb{I}[o'(x) = \texttt{actionable}],
\end{equation}
where $\mathbb{I}[\cdot]$ is the indicator function. This effectively steers the tool toward regions most likely to yield progress.

While \textsc{SPARTA-G} does not involve learning, it still operates within the MDP framework: its “policy” is a deterministic mapping from SPOC-derived state features to actions. This makes it lightweight, fast to deploy, and effective for coarse transformations, though—as we show in Sec.~\ref{sec:expts}—\textsc{SPARTA-L} yields higher performance in tasks requiring finer directional control. 

Together, the two variants illustrate the versatility of spatial progress-based representations, supporting both hand-crafted greedy control and reinforcement learning within the same framework.
\section{Experimental Evaluation}
\label{sec:expts}

\custompar{Manipulation tasks.} 
Spatially-progressing OSCs occur in a variety of real-world domains (cooking, cleaning, arts and crafts, construction, etc.).  We focus
on cooking,
due to its real-world application value, complexity, and wide array of state-changing actions and constituent objects.
We evaluate on 
\stat{three} challenging
OSC tasks:  
  (1)  \textit{Spreading}: 
    spread a coatable substance across the surface of an object (e.g., syrup on a donut, sauce on pizza), resulting in visible changes in color and texture;
    (2) \textit{\Mashing}: crush an object into a purée-like consistency (e.g., banana or avocado), altering its texture from solid to semi-fluid;
    and (3) \textit{Slicing}:
    slice an uncut object into fine, uniformly thin slices (e.g., cucumber or butter). Large, coarse cuts are not considered complete transformations.
All tasks involve significant and irreversible structural and appearance changes, making them challenging for perception, affordance reasoning, and reward design.

\custompar{Objects.}
We evaluate on \stat{10} diverse real-world objects spanning a wide range of shapes, textures, and colors 
(see Table~\ref{tab:main_results}). 
This diversity stresses both sides of the system: the vision model must robustly perceive objects with highly varied appearances, while the control policy must operate across different geometries and material properties.

\custompar{Tool-use primitives.}
To translate high-level policy outputs into physical interactions, we define simple task-specific motion primitives at the predicted 2D location on the object, following prior work using motion primitives for point-based control~\cite{bahl2023vrb,nasiriany2022maple}.
For \textit{spreading}, the robot executes in-plane brush strokes, automatically “refilling” the brush every two steps to keep it coated; \cut{the }$z$-height is fixed relative to the estimated object surface.
For \textit{mashing} and \textit{slicing}, it performs a lateral motion followed by a downward press until a preset force threshold is reached. In all tasks, the tool is lifted after each action to avoid visual occlusion before capturing the next observation.

\custompar{Robot platform.} All experiments are conducted on a Franka Emika Panda robot, a 7-DoF collaborative manipulator equipped with torque sensing in each joint and a 3-finger parallel gripper. Its precise joint control and compliant torque feedback make it well-suited for fine manipulation tasks such as spreading, mashing, and slicing. A front-facing camera provides RGB observations for the vision model.

\custompar{Training and implementation details.} 
To keep training grounded in real-world constraints, we set episode lengths to match the natural granularity of each task. For spreading, episodes last 10 steps to reflect the smaller coverage per action, while for mashing and slicing, 5 steps suffice due to the broader area transformed by each action. All episodes begin from a fixed corner of the workspace for consistency.
For \textsc{SPARTA-L}, we train policies with short real-world budgets: spreading is trained for \stat{40} episodes ($\sim$\stat{3} hours at 1 Hz, including brush refills and resets), while mashing and slicing converge within $\sim$\stat{1.5} hours thanks to shorter episodes. To simplify resets, we use clay proxies for mashing and slicing, and bootstrap exploration with a handful ($\sim$5) greedy rollouts, which stabilize early training.
For policy learning, we adopt asynchronous SAC from SERL~\cite{luo2024serl}, finding that an actor-to-critic update ratio of 1:10 yields the best balance between policy improvement and stable value estimation. Other hyperparameters follow standard practice (learning rate 3e-4 with warmup, $\gamma=0.95$, reward weights $\alpha=1,\beta=1,\eta=0.001$). Visual inputs are encoded via a ResNet-10 backbone, and proprioceptive inputs through a two-layer MLP.
The same training protocol is applied across our method and baselines to ensure fair comparison.

\custompar{Comparisons.} 
We benchmark against three baselines:
\begin{enumerate}[label=(\textbf{\arabic*}),leftmargin=*]
    \item \textsc{Random}: a control baseline with actions sampled uniformly randomly within the constrained action space, reflecting unstructured exploration with no task guidance.
    \item \textsc{Sparse}: a sparse reward baseline using only a binary task completion reward, queried from GPT-4o via task-specific prompts on the final image (e.g., “Is the bread fully coated with ketchup?”). A “yes” ends the episode with +1 reward; otherwise, no reward is given. This mirrors our use of VLMs for SPOC mask generation.
    \item \textsc{LIV}~\cite{ma2023liv}: 
    a state-of-the-art goal-conditioned representation learning method trained on human activity videos~\cite{grauman2022ego4d,Damen2018EPICKITCHENS}. 
    Rewards are computed from state embedding similarities to a language goal.
    We directly prompt LIV with a natural language description of the OSC task and object (e.g., “coat bread with ketchup”).
\end{enumerate}
The baselines represent two dominant approaches: sparse rewards with minimal supervision and pretrained goal-based visual representations. They highlight the limitations of current visual RL methods when applied to fine-grained OSC tasks. We do not include tactile-based or simulation-heavy methods~\cite{heiden2021disect,xu2023roboninja}, as they require task-specific instrumentation. 
Further, unlike imitation learning approaches, SPARTA does not require demonstrations.
Thus, we focus on general, vision-driven approaches requiring no human demonstrations—hence directly comparable to SPARTA.

\custompar{Metric.}
We measure performance using \textit{transformation coverage}: the percentage of the object’s area that has changed state by the end of an episode. Coverage is computed from SPOC segmentations and corrected using~\cite{toras} with human annotators to ensure reliability. Unlike binary success, this continuous metric captures partial progress, providing a more sensitive evaluation of OSC performance.

\subsection*{Experiments and Results:}
\label{subsec:expt_results}

In our experiments, we aim to answer \stat{three} key questions:

\begin{table}[!t]
\centering
\scriptsize
\setlength{\tabcolsep}{0.6pt}
    \begin{tabular}{l | c !{\color{gray!120}\vline} c c c c | c !{\color{gray!120}\vline} c c | c !{\color{gray!120}\vline} c c c c }
    \toprule
     & \multicolumn{5}{c|}{Spread} & \multicolumn{3}{c|}{Slice} & \multicolumn{4}{c}{Mash} \\
     & \multicolumn{1}{c}{\emph{Seen}} & \multicolumn{4}{c|}{\emph{Unseen}} 
     & \multicolumn{1}{c}{\emph{Seen}} & \multicolumn{2}{c|}{\emph{Unseen}} 
     & \multicolumn{1}{c}{\emph{Seen}} & \multicolumn{3}{c}{\emph{Unseen}} \\
    Model
    & \includegraphics[height=2.1em]{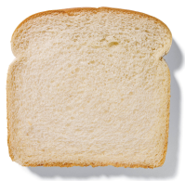}
    & \includegraphics[height=2.1em]{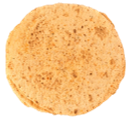}
    & \includegraphics[height=2.1em]{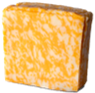}
    & \includegraphics[height=2.1em]{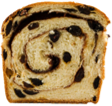}
    & \includegraphics[height=2.3em]{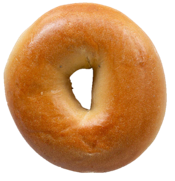}
    & \includegraphics[height=2.4em]{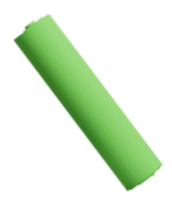} 
    & \includegraphics[height=2.3em]{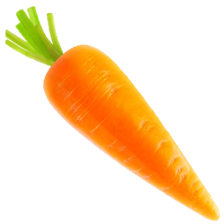}
    & \includegraphics[height=2.4em]{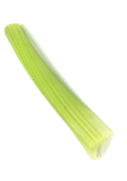}
    & \includegraphics[height=2.4em]{figures/food_icons/clay.png} 
    & \includegraphics[height=2.6em]{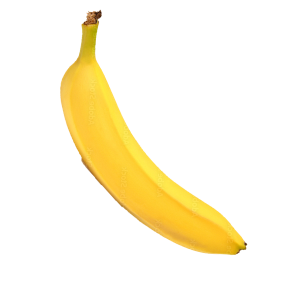} 
    & \includegraphics[height=2.2em]{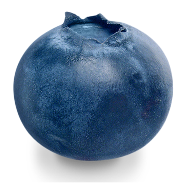} 
    & \includegraphics[height=2.6em]{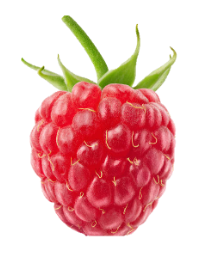} \\
    & \includegraphics[height=1.75em]{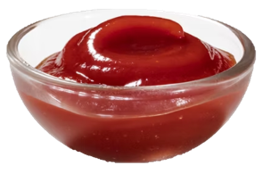}
    & \includegraphics[height=1.72em]{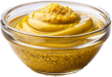}
    & \includegraphics[height=1.75em]{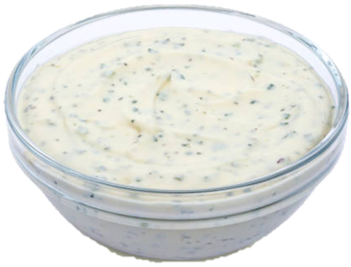}
    & \includegraphics[height=1.75em]{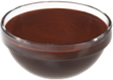}
    & \includegraphics[height=1.75em]{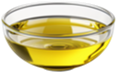} 
    & & & & & & & \\
    \midrule
    \textsc{Random} & 0.24 & 0.42 & 0.27 & 0.29 & 0.23 & 0.13 & 0.15 & 0.14 & 0.18 & 0.14 & 0.23 & 0.20 \\
    \textsc{Sparse} & 0.14 & 0.10 & 0.07 & 0.11 & 0.13 & 0.09 & 0.08 & 0.09 & 0.13 & 0.08 & 0.09 & 0.18 \\
    \textsc{LIV}~[10] & 0.17 & 0.14 & 0.12 & 0.16 & 0.12 & 0.10 & 0.09 & 0.11 & 0.13 & 0.09 & 0.10 & 0.08 \\
    \textbf{\method-G} & 0.44 & 0.49 & 0.55 & \textbf{0.66} & 0.39 & 0.63 & 0.60 & 0.61 & 0.75 & 0.69 & \textbf{0.71} & \textbf{0.75} \\
    \textbf{\method-L} & \textbf{0.61} & \textbf{0.55} & \textbf{0.58} & 0.63 & \textbf{0.42} & \textbf{0.78} & \textbf{0.69} & \textbf{0.67} & \textbf{0.77} & \textbf{0.72} & 0.62 & 0.68 \\
    \bottomrule
    \end{tabular}
\caption{\small{
\method shows strong training and generalization results for objects with varying textures, colors and shapes. 
Metric is transformation coverage (\%). 
Results averaged over 3 seeds, 5 rollouts per seed (15 evaluations total).
}}
\label{tab:main_results}
\vspace*{-0.5cm}
\end{table}

\begin{figure}
    \centering
    \includegraphics[width=\linewidth,trim={0 0.0cm 0 0},clip]{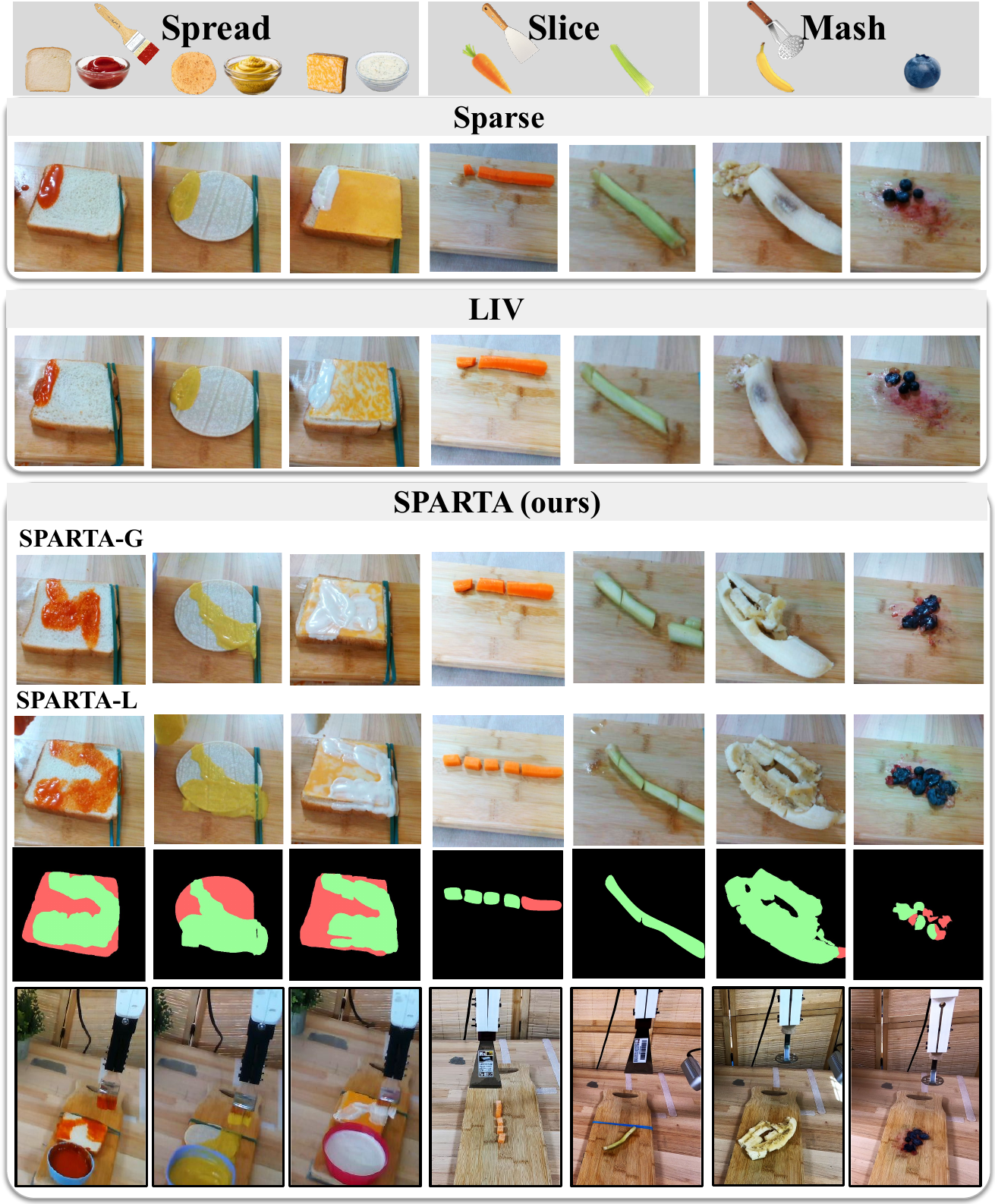}
    \caption{\small{
    SPARTA significantly outmatches the baselines vis-a-vis substantially transforming actionable regions across diverse objects with varying colors, shapes, and textures.
    }}
    \label{fig:qual_results}
    \vspace{-1.5em}
\end{figure}

\custompar{Q1) How well does \method perform complex object state changes?}
Table~\ref{tab:main_results}
presents a comprehensive evaluation of SPARTA across the three tasks spanning both seen and unseen objects, with qualitative results in Fig.~\ref{fig:qual_results}. Across the board, both variants of SPARTA---SPARTA-G (greedy) and SPARTA-L (learning)---dramatically outperform all baselines,
underscoring the power of spatial object-centric affordances in guiding manipulation. \textsc{Sparse} provides little guidance,
while \textsc{LIV}~\cite{ma2023liv}—despite state-of-the-art representation learning—closely follows \textsc{Sparse}, failing
to capture fine-grained spatial progress.
Interestingly, the\cut{ exploratory} \textsc{Random}\cut{ action} policy 
outperforms both baselines, accentuating
how in settings with unstable or weak rewards, RL policy training can lead to degenerate solutions over time due to lack of entropy.

Among our methods, \textsc{SPARTA-G} exploits directional priors in SPOC maps to consistently target actionable regions. It is especially competitive in mashing, where large symmetric tools cover broad areas and dampen the impact of noisy perception. By contrast, \textsc{SPARTA-L} excels in spreading and slicing, where thin asymmetric tools demand precise, noise-robust control. By optimizing long-horizon returns, the learned policy compensates for real-world noise in image-action mappings and achieves pixel-accurate transformations. In short, \textsc{SPARTA-G} suffices when broad, symmetric actions can overcome perceptual noise, whereas \textsc{SPARTA-L} is essential when fine directional control is required.

This comparison underscores a broader takeaway: Spatial OSC segmentation is a powerful and general-purpose visual representation that enables both fast greedy planning and robust reinforcement learning, depending on task complexity.

\begin{figure*} 
    \centering 
    \includegraphics[width=\linewidth]{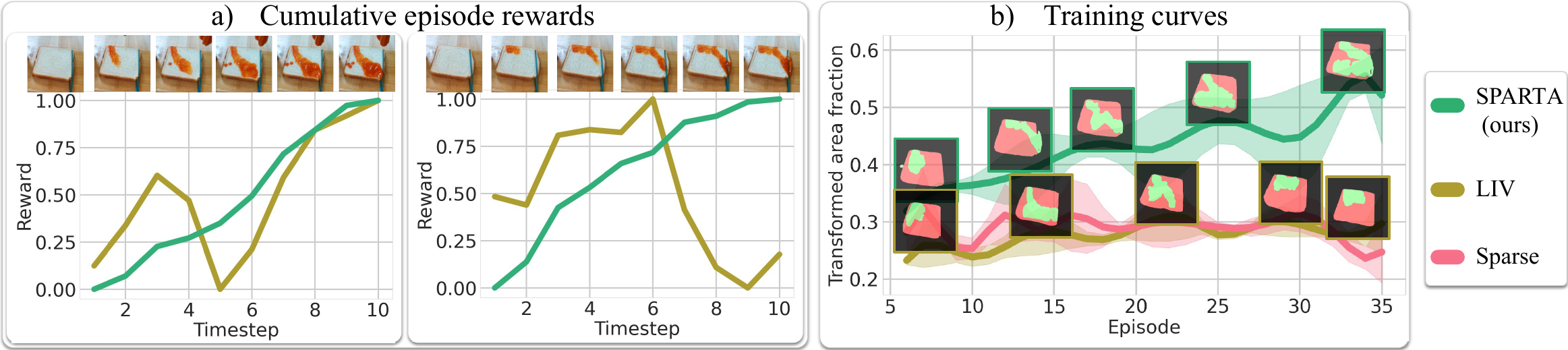} 
    \caption{\small{
    Reward curves for bread-spreading task.
    \textbf{a)} Cumulative episode reward curves: \method produces smooth, incremental rewards aligned with visual progress, while \textsc{LIV} rewards remain unstable throughout the episode, offering poor guidance.
    \textbf{b)} Training curves: stable, dense feedback drives sample-efficient learning, with \method rapidly improving while \textsc{Sparse} and \textsc{LIV} stagnate.
    }}
\vspace*{-0.05in}
\label{fig:reward_curves} 
\end{figure*}

\begin{figure}[!t] 
\centering
    \includegraphics[width=\linewidth]{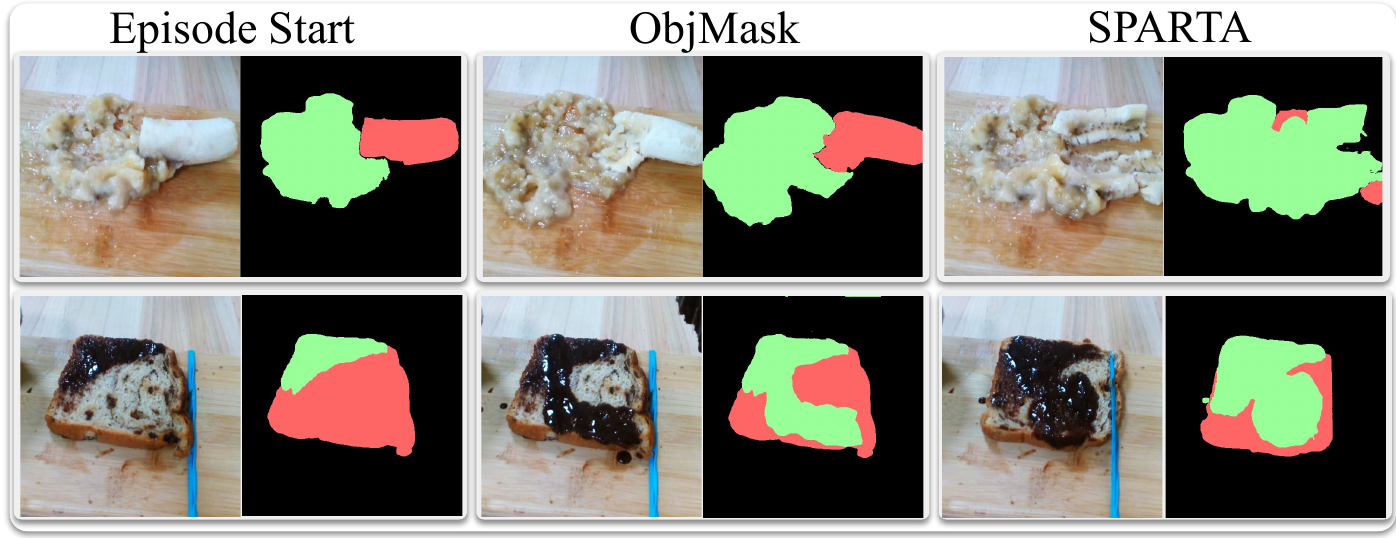} 
    \caption{\small{ 
    Compared to \textsc{ObjMask}, a state-change agnostic policy which naively traverses the entire object area and repeatedly acts on already transformed regions, \textsc{SPARTA} selectively targets only actionable areas, driving efficient state progression.
    }}
    \label{fig:objmask}
\vspace*{-0.1in}
\end{figure}

\custompar{Q2) How do stable per-step episode rewards enable sample-efficient learning?}
A key determinant of real-world sample efficiency is the stability of dense rewards within each episode~\cite{gupta2022unpacking}.
Fig.~\ref{fig:reward_curves}-a shows cumulative reward curves for \method-L and LIV across sample episodes on the bread-spreading task. 
\method produces \textit{smooth, monotonic} curves that align directly with visual progress, leading to consistent incremental rewards. 
In contrast, \textsc{LIV} rewards fluctuate erratically, reflecting how goal-conditioned embeddings fail to capture fine-grained transformation dynamics. 
These unstable signals offer poor guidance, leading to degenerate solutions over time.

This reward stability translates into far more efficient training (Fig.~\ref{fig:reward_curves}-b). 
\method-L exhibits steep, monotonic learning curves from the very first episodes, often reaching usable policies ($>$60\% coverage) within just \stat{90 minutes} of real-world training. 
By contrast, both \textsc{Sparse} and \textsc{LIV} remain flat, unable to improve beyond chance due to the absence of dense, progress-aware feedback. 
Interestingly, the affordance prior also acts as an implicit curriculum: early on, policies focus on small patches of the object, before gradually covering larger regions and learning strategies such as reversing direction near object boundaries. 

Together, these results demonstrate how \method's dense rewards provide stable, interpretable feedback that not only reflects spatial progress but also drives sample-efficient policy learning in the real world.

\custompar{Q3) What is the utility of state change segmentations for robot learning over plain object segmentation maps?}
To isolate the benefit of SPOC affordances over traditional object segmentation, we compare SPARTA-G against a greedy baseline that traverses the entire object segmentation mask without reasoning about state change. We initialize objects in partially transformed states (e.g., a half-mashed banana) and evaluate if the policy can target the remaining untransformed regions (see Fig.~\ref{fig:objmask}). The object segmentation baseline, being agnostic to intra-object state change, wastes actions by repeatedly revisiting already transformed regions (e.g., mashing an already mashed banana segment). In contrast, SPARTA-G, exploits SPOC maps to selectively target only the actionable (untransformed) regions, 
achieving 3$\times$ higher coverage efficiency.
This validates the efficacy of \method for spatially-progressive manipulation policies that reason over state change dynamics—not just object presence.

\section{Lessons and Conclusion}
\label{sec:conclusion}

This work investigates real-world robot learning for a family of spatially-progressing manipulation tasks—such as spreading, mashing, and slicing—by leveraging dense visual affordances that signal object state change. Our method, \method, uses a unified spatial-progress representation to support both greedy planning and reinforcement learning, allowing policy generation for very challenging tasks without simulation or human demonstrations, while generalizing across diverse objects.
While effective, \method also reveals open challenges that suggest avenues for future research.
First, although policies generalize to new objects, performance degrades on unseen geometries—for example, a policy trained on rectangular slices may struggle with circular tortillas. Addressing this gap calls for shape-aware training or augmented experience.
Second, our current pipeline avoids occlusion by only capturing visual inputs when the end-effector lifts between actions, precluding continuous perception during contact. Developing occlusion-resilient, contact-aware visual reasoning remains an open challenge. 
Overall, \method demonstrates that progress-aware affordances can unlock a family of object state manipulations essential for everyday tasks, charting a path beyond rigid-body control.




\section*{ACKNOWLEDGMENT}
Work supported in part by UT Austin IFML NSF AI Institute and DARPA TIAMAT program (HR0011-24-9-0428).
We thank Luca Macesanu for designing the tool attachments.

\printbibliography

\end{document}